\documentclass[conference]{IEEEtran}
\IEEEoverridecommandlockouts

\usepackage[utf8]{inputenc} 
\usepackage[T1]{fontenc}
\usepackage{url}
\usepackage{ifthen}
\usepackage{cite}
\usepackage{bm}
\usepackage[cmex10]{amsmath} 
\usepackage{amssymb}
\DeclareMathOperator*{\argmax}{arg\,max}

\usepackage{xcolor}
\newtheorem{lemma}{Lemma}
\usepackage{graphicx}
\usepackage{dblfloatfix}

\begin{document}
\title{Queueing-Aware Optimization of Reasoning Tokens for Accuracy–Latency Trade-offs in LLM Servers 
\thanks{This work was supported by Tubitak 2232-B program (Project No:124C533).}
} 

\author{
  \IEEEauthorblockN{Emre Ozbas and Melih Bastopcu}
 \IEEEauthorblockA{Department of Electrical and Electronics Engineering\\
                    Bilkent University, Ankara, Türkiye\\
                    Email: \{emre.ozbas, bastopcu\}@bilkent.edu.tr} }


\maketitle


\begin{abstract}
We consider a single large language model (LLM) server that serves a heterogeneous stream of queries belonging to $N$ distinct task types. Queries arrive according to a Poisson process, and each type occurs with a known prior probability. For each task type, the server allocates a fixed number of internal thinking tokens, which determines the computational effort devoted to that query. The token allocation induces an accuracy-latency trade-off: the service time follows an approximately affine function of the allocated tokens, while the probability of a correct response exhibits diminishing returns. Under a first-in, first-out (FIFO) service discipline, the system operates as an $\mathrm{M}/\mathrm{G}/1$ queue, and the mean system time depends on the first and second moments of the resulting service-time distribution. We formulate a constrained optimization problem that maximizes a weighted average accuracy objective penalized by the mean system time, subject to architectural token-budget constraints and queue-stability conditions. The objective function is shown to be strictly concave over the stability region, which ensures existence and uniqueness of the optimal token allocation. The first-order optimality conditions yield a coupled projected fixed-point characterization of the optimum, together with an iterative solution and an explicit sufficient condition for contraction. Moreover, a projected gradient method with a computable global step-size bound is developed to guarantee convergence beyond the contractive regime. Finally, integer-valued token allocations are attained via rounding of the continuous solution, and the resulting performance loss is evaluated  in simulation results.
\end{abstract}

\section{Introduction}

Large language models (LLMs) are increasingly deployed as inference services that
must handle many concurrent user requests under strict latency constraints
\cite{kwon2023efficient, low_latency_LLM}. At the same time, a large fraction of
queries require multi-step reasoning, where insufficient computational effort
can lead to incorrect or low-quality outputs. This creates an inherent trade-off
between response accuracy and delay, which has recently attracted significant
attention in the context of LLM serving systems and inference-time control
\cite{S1_STTS, schuster2022confident, pope2023efficiently}. A common approach to
regulating inference-time computation is to limit the amount of internal
reasoning performed by the model, typically through a
\emph{reasoning-token} budget \cite{S1_STTS, wei2022chain}. Increasing this budget
generally improves accuracy but also increases inference time
\cite{pope2023efficiently}, with diminishing returns observed beyond a certain
point \cite{kaplan2020scaling}. When requests are processed sequentially by a
single model instance, longer inference times not only affect individual queries
but also propagate through queueing delays, degrading overall system
responsiveness.

Given the high latency of LLM inference, a wide range of techniques have been
proposed to reduce inference time at different levels of the overall system.
At the model level, reference \cite{liu2023contextual} 
demonstrates the existence of
contextual sparsity in LLMs, enabling inference to activate only a subset of
parameters conditioned on the input, rather than executing the full model. From a systems perspective, \cite{zheng2023microbatch} 
proposes predicting response lengths and grouping queries with similar generation
profiles into micro-batches to improve throughput efficiency. Hardware-aware approaches have also been explored,
including INT8 quantization of weights and activations for matrix
multiplications in LLMs \cite{xiao2023int8}. While these methods are effective at
reducing per-request inference time, they primarily target execution efficiency
and do not explicitly account for stochastic arrivals or queueing delays in LLM
serving systems.

\begin{figure}[t]
\centering
\includegraphics[width=0.95\columnwidth]{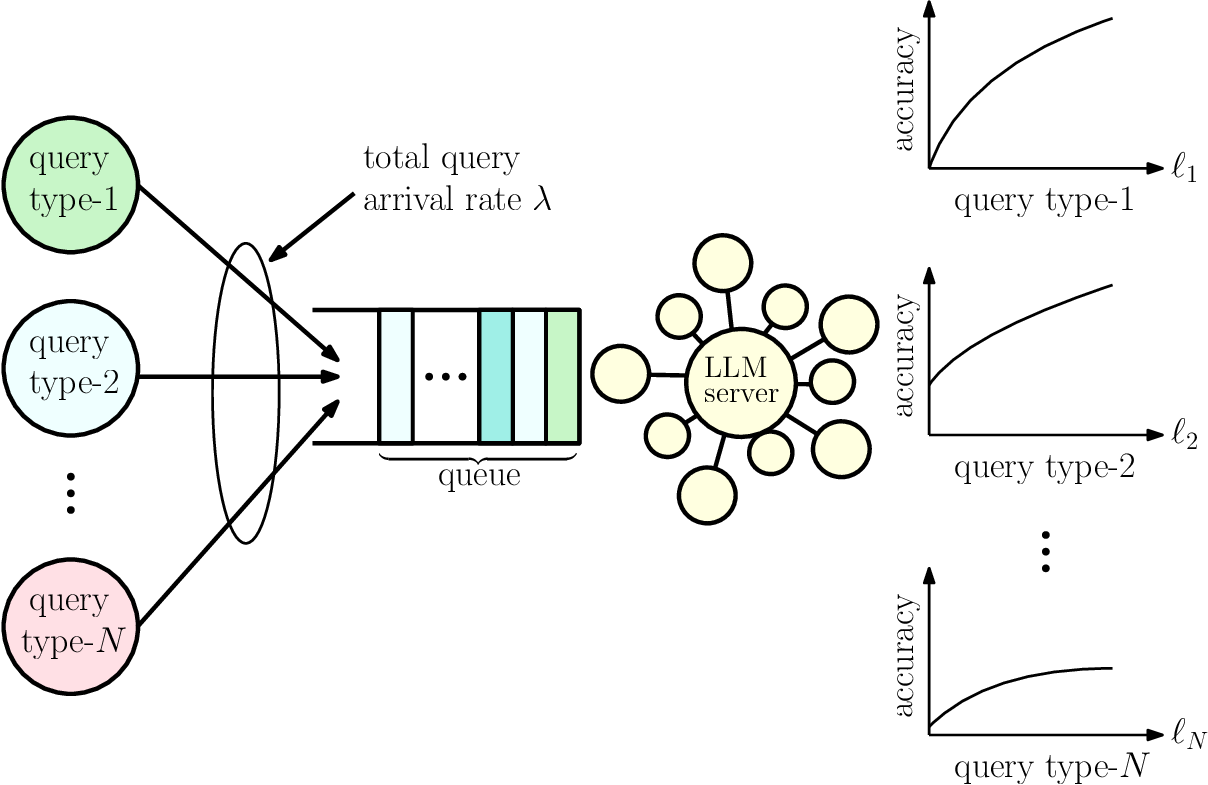}
\vspace{-0.35cm}
\caption{System model of a single LLM server processing $N$ different heterogeneous query
types.}
\label{fig:system_model}
\vspace{-0.6cm}
\end{figure}

In practice, user workloads are highly heterogeneous: different query types
benefit unequally from additional reasoning effort
\cite{schuster2022confident}. Applying a uniform reasoning-token budget across all
queries can therefore be inefficient, as some tasks achieve near-optimal accuracy
with modest computation, while others require substantially larger budgets to
reach comparable performance. This heterogeneity motivates adaptive allocation
of reasoning resources across query types, accounting for both accuracy gains and
their impact on system-level latency.

Recently, queueing-theoretic approaches have begun to provide insights into the
latency behavior of LLM inference services. \cite{li2023parallel} 
analyzes LLM
serving from a parallelization perspective, characterizing how multiple workers
can reduce response times under stochastic arrivals. 
Most closely related to our work, reference \cite{low_latency_LLM} studies LLM inference using a
queueing-theoretic framework that explicitly models variability in \emph{output token
lengths.} They represent the inference system as an M/G/1
queue, where the service-time distribution is governed by response length, and
show that imposing a maximum output token limit can substantially reduce
system-wide queueing delays while affecting only a small fraction of long
responses. Their results underscore the critical role of token-length variability
in shaping latency and illustrate how inference-time control mechanisms can be
used to improve responsiveness in LLM serving systems. While \cite{low_latency_LLM} controls latency via a \emph{global} limit on
\emph{output} tokens, our control variable is the \emph{per-task reasoning-token
budget}. Specifically, we model a heterogeneous workload with $N$ task types
and assign a task-dependent reasoning length $\ell_k$ that shapes both accuracy and service time. This
yields a coupled accuracy-queueing delay trade-off and enables a systematic token-allocation optimization across tasks.

Motivated by these observations, we study a single LLM server, as illustrated in
Fig.~\ref{fig:system_model}, that processes a stream of heterogeneous queries and
investigate how per-type reasoning-token budgets should be selected to balance
response accuracy and latency. We formulate a system-level performance objective
that jointly captures average accuracy and mean system time, subject to
architectural constraints and stability conditions \cite{bertsekas1987data}. We
show that this objective is strictly concave over the feasible region, ensuring
existence and uniqueness of the optimal token allocation. Exploiting this
structure, we derive a projected fixed-point iteration based on the
Karush-Kuhn-Tucker (KKT) conditions and establish explicit sufficient conditions
for its convergence. To ensure robustness beyond the contractive regime, we also
propose a projected gradient ascent method with a computable global step-size
bound. Finally, we obtain integer-valued token lengths through rounding and
quantify the resulting performance loss via simulation results.

\section{System Model and Problem Formulation} \label{sec:system_model}
In this work, we consider a single LLM server that processes a stream
of incoming user queries. Query arrivals follow a Poisson process with rate $\lambda>0$. Each query belongs to one of $N$ task categories indexed by
$k \in \{1,\dots,N\}$, occurring with probabilities $\pi_k$'s satisfying
$\sum_{k=1}^N \pi_k = 1$. Each query type is realized independently from each other. Thus, the query arrival process for task category $k$ follows  a Poisson process with rate $\lambda_k$ given by $\lambda_k = \pi_k \lambda$. These categories can correspond to heterogeneous
reasoning tasks, such as solving mathematical problems or determining the output of a given code snippet. A type-$k$ query is assigned a deterministic reasoning-token length
$\ell_k \ge 0$. This quantity specifies the number of internal reasoning tokens
generated by the model. A strict budget-enforcement mechanism ensures that
exactly $\ell_k$ tokens are produced \cite{S1_STTS}. The budgets are subject to
the architectural constraint $0 \le \ell_k \le \ell_{\max}$ for all $k$
where $\ell_{\max}$ reflects the memory and structural limitations of the LLM that bound the maximum amount of reliable computation.

Empirical latency measurements demonstrate that
the inference time scales approximately linearly with the number of reasoning
tokens \cite{low_latency_LLM}. Accordingly, we model the service time of a type-$k$ query as
\begin{align}\label{eqn:reasoning_time}
t_k(\ell_k) = t_{0k} + c_k \ell_k,
\end{align}
where $t_{0k}$ represents the fixed overhead associated with processing the input prompt (prefill phase) and loading model weights and $c_k$ denotes the per-token
processing time.

\begin{figure}[t]
\centering
\includegraphics[width=\columnwidth]{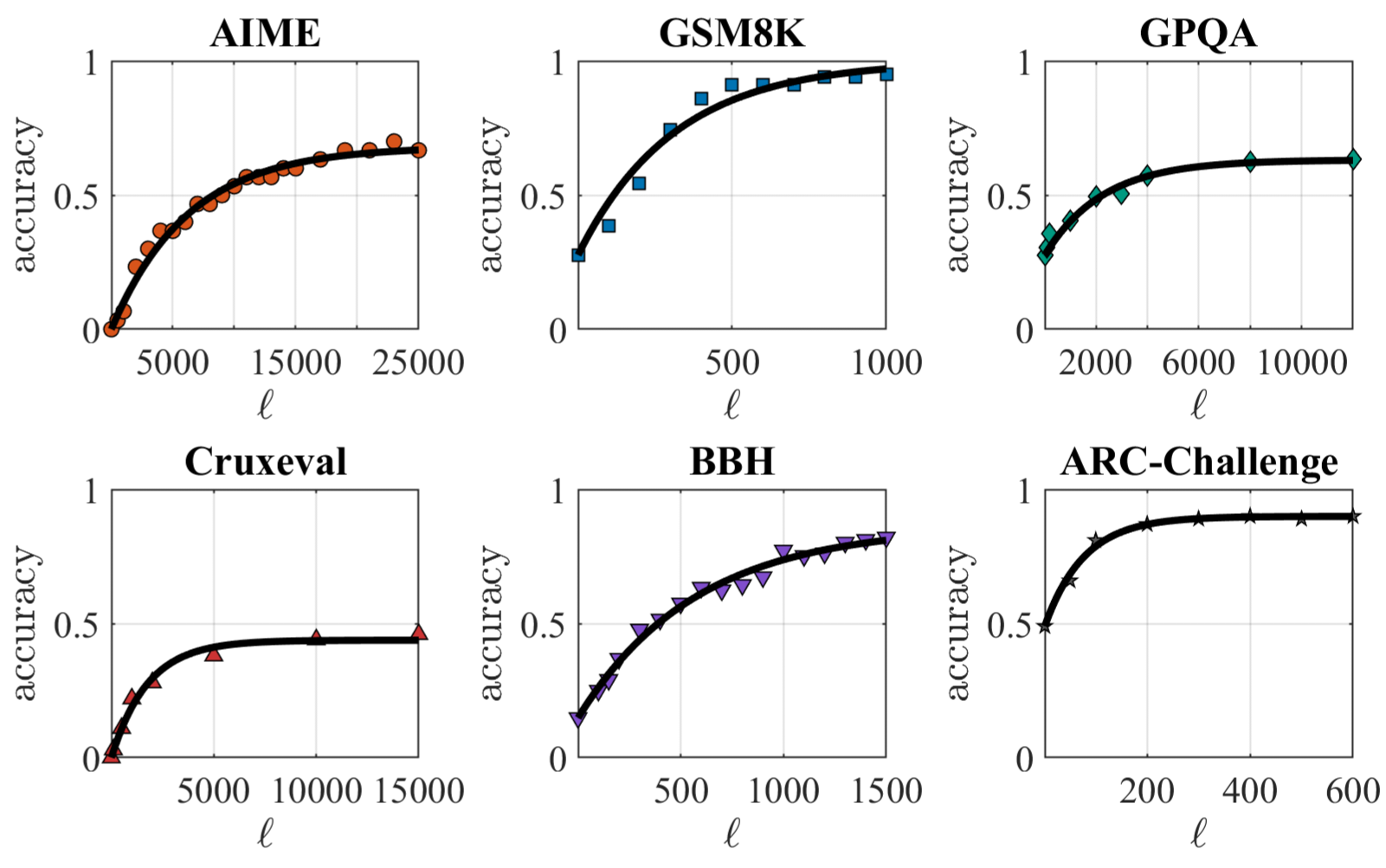}
\vspace{-0.7cm}
\caption{Empirical accuracy as a function of the enforced reasoning-token budget $\ell$ for each task type. Markers denote measured accuracies; solid curves show the fitted model $p_k(\ell)=A_k(1-e^{-b_k\ell})+D_k$.}
\label{fig:accuracy_curve}
\vspace{-0.5cm}
\end{figure}

Similarly, the empirical accuracy curves as shown in Fig.~\ref{fig:accuracy_curve} (which we will discuss in more detail in Section~\ref{sec:numerical_results})  exhibit diminishing returns in accuracy with the number of reasoning tokens.
To capture this behavior, the probability of an accurate response for a type-$k$ query is modeled as
\begin{align}
p_k(\ell_k)
=
A_k \left(1 - e^{-b_k \ell_k}\right) + D_{k}, 
\end{align}
where $A_k \in (0,1]$ and $D_k \in (0,1]$. Here, $A_k +D_k$ is the maximum attainable accuracy such that $A_k +D_k\leq 1$ and $b_k > 0$
controls the curvature of the accuracy-reasoning tokens relationship. Next, we present the queueing dynamics for the system model described above.

\subsection{Queueing Dynamics}

Under the LLM server model described above, the system allocates  $\ell_k$ reasoning tokens to each query of type $k$. As a result, the LLM server spends a constant processing time $t_k(\ell_k)$ in (\ref{eqn:reasoning_time}). Then, the overall service time for the queries is a random variable $S$ which takes the value $t_k(\ell_k)$ with probability $\pi_k$. The first and the second moments of random variable $S$ are given by:    
\begin{align}
\mathbb{E}[S] = \sum_{k=1}^N \pi_k\, t_k(\ell_k), \quad \mathbb{E}[S^2] = \sum_{k=1}^N \pi_k\, t_k(\ell_k)^2.\label{eqn:service:first_moment}
\end{align}
Upon the arrival of a query, if the LLM server is idle, it immediately begins processing the query. If the server is already busy, the new query waits in a queue, where all queries are served in their order of arrival, following the first-in, first-out (FIFO) discipline. As a result, the LLM server operates as an $\mathrm{M}/\mathrm{G}/1$ queue where the inter-query arrival times are exponentially distributed with rate $\lambda$, and the service time $S$ follows a general distribution whose first and second moments, i.e., $\mathbb{E}[S]$ and $\mathbb{E}[S^2]$, are provided in (\ref{eqn:service:first_moment}). The resulting $\mathrm{M}/\mathrm{G}/1$ queue is stable if and only if \cite{bertsekas1987data}
\begin{equation}
\lambda\, \mathbb{E}[S] < 1.
\end{equation}
Under this stability condition, the mean waiting time in the queue follows from
the Pollaczek-Khinchine formula \cite{bertsekas1987data}:
\begin{align}
\mathbb{E}[W]
=
\frac{\lambda\, \mathbb{E}[S^2]}
     {2\bigl(1 - \lambda\, \mathbb{E}[S]\bigr)}.
\end{align}
Then, the mean system time of a query (including both its waiting time in the queue and its service time) is given by
\begin{equation}
\mathbb{E}[T_{\mathrm{sys}}]
=
\mathbb{E}[W] + \mathbb{E}[S].
\end{equation}

These expressions show how the reasoning-token length $\ell_k$'s jointly shape both the accuracy and the latency of the system, giving rise to an accuracy-delay trade-off.

\subsection{Problem Formulation}

We now formalize the joint accuracy-latency trade-off induced by the reasoning-token lengths $\ell_k$'s. Our objective is to determine the optimal values of $\ell_k$'s that maximize the query response accuracy while minimizing the system time of the LLM server. Thus, we aim to maximize the overall performance objective of the system, defined as:
\begin{equation}
J(\bm{\ell})
= \alpha \sum_{k=1}^N \pi_k\, p_k(\ell_k)-
\frac{\lambda\, \mathbb{E}[S^2]}
     {2\left(1 - \lambda\, \mathbb{E}[S]\right)}
- \mathbb{E}[S],
\label{eq:objective}
\end{equation}
where $\alpha > 0$ controls the relative importance assigned to the response accuracy and $\bm{\ell} = \{\ell_1,\cdots,\ell_N\}$. The first term ($\sum_{k=1}^N \pi_k\, p_k(\ell_k)$) quantifies the average accuracy gains achieved through increased reasoning tokens and thus longer processing times while the remaining two terms represent the mean system time.

In this problem, the reasoning tokens are discrete, so for each query type, we have 
\begin{equation}
\ell_k \in \{0,1,\cdots,\ell_{\max}\}, \qquad k = 1,\dots,N.
\label{eq:integer_tokens}
\end{equation}
Such an optimization over discrete variables results in a mixed-integer optimization problem, which is generally computationally hard. To obtain a tractable formulation, we first consider the standard continuous relaxation in which each $\ell_k$ is allowed to take any real value in the interval of $[0,\ell_{\max}]$. All structural results and algorithms developed in
the following sections are derived for this relaxed problem. The final implementable integer policy is then obtained by
projecting the optimal continuous solution onto the discrete set
\eqref{eq:integer_tokens} as we will demonstrate explicitly in Subsection~\ref{sec:integer_impl}.

In the relaxed problem, the feasible set is determined by models architectural limits and the queue stability condition. Thus, the resulting (relaxed) optimization problem is  given by
\begin{align}
\label{eq:opt_problem}
\max_{\bm{\ell} \in \mathbb{R}^N}  \quad & J(\bm{\ell}) \nonumber \\
\mbox{s.t.} \quad & 0 \le \ell_k \le \ell_{\max}, \quad k = 1,\dots,N, \nonumber \\
\quad &  \lambda\, \mathbb{E}[S(\ell)] < 1.
\end{align}

Solving this relaxed problem yields the optimal set of reasoning-token lengths that achieve the optimal balance between response accuracy and latency. The discrete token allocation actually implemented by the LLM server is then obtained by
rounding these continuous reasoning-token lengths to feasible integers.

\section{Optimal Token Allocation}
\label{sec:optimal_allocation}

In this section, we characterize the structure of the optimal reasoning-token lengths
$\boldsymbol{\ell} = \{\ell_k\}_{k=1}^N$ that solve~\eqref{eq:opt_problem}. We first show that the objective function in \eqref{eq:opt_problem} is concave and the feasible set is convex which makes the overall optimization problem convex, guaranteeing that any solution satisfying the first-order optimality conditions is globally optimal. We then write the associated Lagrangian and
state the KKT conditions that characterize the optimum solution.

\subsection{The Convexity of the Optimization Problem}

We first show that the objective $J(\bm{\ell})$ in~\eqref{eq:objective} is concave in
the vector of token budgets $\bm{\ell} = (\ell_1,\cdots,\ell_N)$.

\begin{lemma}
\label{lem:convexity}
On the stability region $\{\bm{\ell} : \lambda\,\mathbb{E}[S(\bm{\ell})] < 1\}$,
the objective $J(\bm{\ell})$ is strictly concave.
\end{lemma}

\begin{IEEEproof}
The derivative of the expected waiting time for a query with respect to the reasoning-token length $\ell_k$, that is $\frac{\partial \mathbb{E}[W]}{\partial \ell_k }$, is given by
\begin{align}\label{eqn:partial_der}
   \frac{\partial \mathbb{E}[W]}{\partial \ell_k } = \lambda \pi_k c_k \left(\frac{t_k(\ell_k)}{1-\lambda \mathbb{E}[S]} + \frac{\lambda \mathbb{E}[S^2]}{2(1-\lambda \mathbb{E}[S])^2}\right). 
\end{align}
Similarly, we find the second derivative of the expected waiting time for a query as

\begin{align}\label{eqn:partial_der2}
  \!  \frac{\partial^2 \mathbb{E}[W]}{\partial \ell_k \partial \ell_j } \!\!=\!\! \begin{cases} 
      \!\frac{\lambda \pi_k c_k^2 }{1\!-\!\lambda \mathbb{E}[S]}  \left(\!\frac{2\lambda \pi_k t_k(\ell_k)}{1\!-\!\lambda \mathbb{E}[S]} \!+\! \frac{\lambda^2 \pi_k \mathbb{E}[S^2]}{(1\!-\!\lambda \mathbb{E}[S])^2} \!+ \!1\!\right), \!\! & \text{if } k=j \\
       \\
      \frac{\lambda^2\pi_k c_k\pi_j c_j}{(1\!-\!\lambda \mathbb{E}[S])^2}\left(t_k(\ell_k)\!+\!t_j(\ell_j)\!+\!\frac{\lambda\mathbb{E}[S^2] }{1\!-\!\lambda \mathbb{E}[S]}\right),
      & \text{if } k\neq j 
   \end{cases}
\end{align}
Then, we form the Hessian matrix $\bm{H}\in\mathbb{R}^{N\times N}$ comprised of the diagonal elements given as $\frac{\partial^2 \mathbb{E}[W]}{\partial \ell_k^2 }$ and off-diagonal elements given by $\frac{\partial^2 \mathbb{E}[W]}{\partial \ell_k \partial \ell_j }$. To prove that $\mathbb{E}[W]$ is strictly  convex with respect to the vector $\bm{\ell}$, we need to show that the Hessian matrix $\bm{H}$ is positive definite. In other words, for any real valued vector with $\bm{z}\neq 0$ where $\bm{z}=[z_1,\dots,z_N]^T \in \mathbb{R}^N$, we must have $\bm{z}^T \bm{H}\bm{z}>0 $. We find $\bm{z}^T \bm{H}\bm{z}$ as
\begin{align}\label{eqn:pos_def_1}
  \bm{z}^T \bm{H}\bm{z} = \frac{\lambda^3\mathbb{E}[S^2]}{D^3} B^2+\frac{2\lambda^2}{D^2}AB+\frac{\lambda}{D}\sum_{k=1}^{N}\pi_k c_k^2 z_k^2 , 
\end{align}
with $A =\sum_{k=1}^{N} \pi_k c_k z_k t_k(\ell_k)$, $B =\sum_{k=1}^{N} \pi_k c_k z_k$, and $D = 1- \lambda \mathbb{E}[S]$.
After some algebraic manipulations, we can rewrite (\ref{eqn:pos_def_1}) as
\begin{align}
    \bm{z}^T \bm{H}\bm{z}\! =\! \frac{\lambda^3\mathbb{E}[S^2]}{D^3}\! \left(\!B\!+\!\frac{DA}{\lambda\mathbb{E}[S^2]}\right)^2\!\!\!+\! \frac{\lambda}{D}\left(\!\sum_{k=1}^{N}\pi_k c_k^2 z_k^2\!-\!\frac{ A^2}{\mathbb{E}[S^2]} \right) 
\end{align}
The equation above is positive if $\sum_{k=1}^{N}\pi_k c_k^2 z_k^2-\frac{ A^2}{\mathbb{E}[S^2]}\geq0$ which implies that 
\begin{align}\label{eqn:pos_cond}
    \!\!\left(\sum_{k=1}^{N} \pi_k c_k z_k t_k(\ell_k)\!\right)^2\!\!\!\leq \!\left(\sum_{k=1}^{N}\pi_k c_k^2 z_k^2\right)\!\left(\sum_{k=1}^{N}\pi_k t_k(\ell_k)^2\!\right)\!.\!\!
\end{align}
It is easy to see that (\ref{eqn:pos_cond}) is true due to Cauchy–Schwarz inequality. As a result, we have $\bm{z}^T \bm{H}\bm{z}>0$ unless $\bm{z}\neq0$ which implies that the Hessian matrix $\bm{H}$ is positive definite. Therefore, $\mathbb{E}[W]$ is strictly convex with respect to the vector $\bm{\ell}$. Equivalently, $-\mathbb{E}[W]$ is a strictly concave function with respect to $\bm{\ell}$. For the accuracy term $p_k(\ell_k)$ in (\ref{eq:objective}), we have
\begin{align}
   \!\! \!\!\frac{\partial p_k(\ell_k) }{\partial \ell_k} \!\!=\!\! A_k b_k e^{-b_k\ell_k}\!>\!0, ~ \frac{\partial^2 p_k(\ell_k) }{\partial \ell_k^2}\! \!=\!\! -A_k b_k^2 e^{-b_k\ell_k}\!<\!0.\!\!
\end{align}
Thus, the accuracy term  $p_k(\ell_k)$ is an increasing concave function of $\ell_k$. As a result, the term $\alpha \sum_{k=1}^N \pi_k\, p_k(\ell_k)$ in $J(\bm{\ell})$ becomes concave. Since the mean service time $\mathbb{E}[S(\bm{\ell})]$ is affine in $\bm{\ell}$, the objective function $J(\bm{\ell})$ is strictly concave.
\end{IEEEproof}
Since $J(\bm{\ell})$ is strictly concave in $\bm{\ell}$ and the feasible set defined by $ 0 \le \ell_k \le \ell_{\max}$ for all $k = 1,\dots,N$ and $  \lambda\, \mathbb{E}[S(\ell)] < 1$ is a convex set, the optimization problem $\max_{\bm{\ell} \in \mathbb{R}^N}  J(\bm{\ell})$ given in (\ref{eq:opt_problem}) which is equivalent to $\min_{\bm{\ell} \in \mathbb{R}^N}  -J(\bm{\ell})$ is convex. 

\subsection{Fixed-Point Iteration via Lagrangian Formulation}

Next, we define the Lagrangian function \cite{boyd2004convex} for the equivalent minimization problem, that is $\min_{\bm{\ell} \in \mathbb{R}^N}  -J(\bm{\ell})$, with the feasibility constraints in (\ref{eq:opt_problem}) as
\begin{align}
\mathcal{L}(\boldsymbol{\ell}, \boldsymbol{\mu}, \boldsymbol{\nu}, \gamma)
&=
-J(\boldsymbol{\ell})
- \sum_{k=1}^{N} \mu_k \ell_k
+ \sum_{k=1}^{N} \nu_k (\ell_k - \ell_{\max})
\nonumber\\ &\quad +
\gamma\bigl(\lambda\,\mathbb{E}[S(\boldsymbol{\ell})]-1\bigr),
\end{align}
where $\mu_k\ge 0$, $\nu_k\ge 0$, and $\gamma\ge 0$ are the Lagrange multipliers.
Next, we state the KKT conditions as follows:
\begin{align}
\frac{\partial \mathcal{L}}{\partial \ell_k} =&
\lambda \pi_k c_k \left(\frac{t_k(\ell_k)}{1-\lambda \mathbb{E}[S]} + \frac{\lambda \mathbb{E}[S^2]}{2(1-\lambda \mathbb{E}[S])^2}\right) +\pi_k c_k \nonumber \\
& -\alpha \pi_k A_k b_k e^{-b_k \ell_k} + \gamma\lambda \pi_k c_k
- \mu_k + \nu_k
= 0 .
\label{eq:kkt_stationarity_final}
\end{align}
Then,  we write the complementary slackness (C.S.) conditions
\begin{equation}
\!\gamma\bigl(\lambda\mathbb{E}[S(\boldsymbol{\ell})]-1\bigr)=0, \quad \mu_k \ell_k = 0,
\quad
\nu_k(\ell_k - \ell_{\max}) = 0,
\label{eq:slackness_final}
\end{equation}
for all $k =1,\cdots, N$.

Since $\lim_{\lambda\mathbb{E}[S(\boldsymbol{\ell})]\rightarrow1}J(\bm{\ell}) =-\infty $, in the optimum solution,
we must have $\lambda\mathbb{E}[S(\boldsymbol{\ell})]<1$ which implies that $\gamma^\star = 0$
from the C.S. conditions in (\ref{eq:slackness_final}). Similarly, by C.S. conditions,
either $\mu_k = 0$ with $\ell_k \ge 0$, or $\ell_k = 0$ with $\mu_k \ge 0$. Moreover,
from $\nu_k(\ell_k - \ell_{\max}) = 0$, it follows that either $\nu_k = 0$ and
$\ell_k \le \ell_{\max}$, or $\nu_k > 0$ and $\ell_k = \ell_{\max}$.

For ease of notation, we define two functions $L_k(\boldsymbol{\ell})>0$ and
$K_k(\boldsymbol{\ell})$ such that the KKT condition in
\eqref{eq:kkt_stationarity_final} can be equivalently written as
\begin{align}
\ell_k - L_k(\boldsymbol{\ell})\, e^{-b_k \ell_k} = K_k(\boldsymbol{\ell}),
\label{eq:LK_simple}
\end{align}
where $L_k(\boldsymbol{\ell})$ and $K_k(\boldsymbol{\ell})$ depend on
$\boldsymbol{\ell}$ through the aggregate moments
$\mathbb{E}[S(\boldsymbol{\ell})]$ and $\mathbb{E}[S^2(\boldsymbol{\ell})]$ and are given by 
\begin{align}
L_k(\boldsymbol{\ell})
&\triangleq
\frac{\alpha A_k b_k}{\lambda c_k^2}
\bigl(1-\lambda\mathbb{E}[S(\boldsymbol{\ell})]\bigr),
\\
K_k(\boldsymbol{\ell})
&\triangleq
-\frac{t_{0k}}{c_k}
-\frac{1\!-\!\lambda\mathbb{E}[S(\boldsymbol{\ell})]}{\lambda c_k}
-\frac{\lambda\,\mathbb{E}[S^2(\boldsymbol{\ell})]}
{2c_k\bigl(1\!-\!\lambda\mathbb{E}[S(\boldsymbol{\ell})]\bigr)}.
\end{align}

We note that $L_k(\boldsymbol{\ell})$ and $K_k(\boldsymbol{\ell})$ depend on
$\ell_j$ for $j = 1,\cdots,N$. This coupling induces a \textit{general fixed-point
mapping} implied by the KKT system. Solving \eqref{eq:LK_simple} for $\ell_k$ yields a closed-form update in terms of
the Lambert-$W$ function \cite{corless1996lambertw}:
\begin{equation}
\widehat{\ell}_k(\boldsymbol{\ell})
=
\frac{1}{b_k}\,
W\!\left(
b_k L_k(\boldsymbol{\ell})\, e^{-b_k K_k(\boldsymbol{\ell})}
\right)
+
K_k(\boldsymbol{\ell}),
\label{eq:LK_lambert_fp}
\end{equation}
where $W(\cdot)$ denotes the principal branch of the Lambert-$W$ function.

Taking into account the feasibility constraints $0 \le \ell_k \le \ell_{\max}$, we can
write the optimal reasoning-token lengths $\ell_k^\star$ as
\begin{equation}
\ell_k^\star
=
\min\!\Bigl\{
\max\!\bigl(\widehat{\ell}_k(\boldsymbol{\ell}^\star),\,0\bigr),
\ \ell_{\max}
\Bigr\},
\quad k=1,\cdots,N.
\label{eq:projected_fp_final_LK}
\end{equation}

\vspace{4pt}
\noindent\textbf{Fixed-point algorithm.}
Equation~\eqref{eq:projected_fp_final_LK} suggests the iterative update given by
\begin{equation}
\!\!\ell_k^{(n+1)}
\!\!=\!
\min\!\Bigl\{
\max\!\bigl(\widehat{\ell}_k(\boldsymbol{\ell}^{(n)}),\,0\bigr),
\ \ell_{\max}
\Bigr\},\!\!
\quad k=1,\cdots,N.\!
\label{eq:proj_fp_iteration_LK}
\end{equation}
Under a contraction condition on the Jacobian of
$\widehat{\boldsymbol{\ell}}(\cdot)$ (which will be discussed in the next subsection),
this projected fixed-point method converges to the unique KKT point
$\boldsymbol{\ell}^\star$.

\subsection{Sufficient Condition for Convergence}
\label{sec:convergence_fp}

Recall the unconstrained fixed-point map
$\widehat{\boldsymbol{\ell}}(\boldsymbol{\ell})$ defined componentwise in
\eqref{eq:LK_lambert_fp} and the projected iteration defined in
\eqref{eq:proj_fp_iteration_LK}. Since the projection onto
$[0,\ell_{\max}]^N$ is nonexpansive, it suffices to establish that the
unprojected mapping $\widehat{\boldsymbol{\ell}}(\cdot)$ is a contraction on
$[0,\ell_{\max}]^N$. For that, we derive an explicit Lipschitz constant based on the Jacobian of $\widehat{\boldsymbol{\ell}}(\cdot)$.

\begin{lemma}
\label{lem:contraction_bound_LW}
We define $t_k^{\max}\triangleq t_{0k}+c_k\ell_{\max}$,
$\mathbb{E}[S]_{\max}\triangleq\sum_{i=1}^N\pi_i t_i^{\max}$, and
$\mathbb{E}[S^2]_{\max}\triangleq\sum_{i=1}^N\pi_i (t_i^{\max})^2$ and assume that
$\rho_{\max}\triangleq \lambda\mathbb{E}[S]_{\max}<1$. Then, for all
$\boldsymbol{\ell}\in[0,\ell_{\max}]^N$ and all $k,j\in\{1,\dots,N\}$, we have
\begin{align}
\left|\!
\frac{\partial \widehat{\ell}_k(\boldsymbol{\ell})}{\partial \ell_j}
\!\right|
&\!\le\!
\frac{\pi_j c_j}{c_k}\!\left[\!
1
\!+\!
\frac{\lambda\, t_{\max}}{1\!-\!\rho_{\max}}
\!+\!
\frac{\lambda^2\,\mathbb{E}[S^2]_{\max}}{2(1\!-\!\rho_{\max})^2}
\!\right]
\!+\!
\frac{\lambda\,\pi_j c_j}{b_k(1\!-\!\rho_{\max})}\!,
\label{eq:LW_partial_bound_compact}\\[-24pt]\nonumber
\end{align}
where $t_{\max}\triangleq\max_i t_i^{\max}$. Hence, we have
\begin{align}\label{eq:L_infty_LW_compact}
\bigl\|\mathcal{J}_{\widehat{\boldsymbol{\ell}}}(\boldsymbol{\ell})\bigr\|_\infty
\!\le\!
L_\infty
\!\triangleq\!&
\max_k\!\left\{\!
\frac{1}{c_k}\!\left[\!
1\!+\!\lambda\!\left(\!
\frac{t_{\max}}{1\!-\!\rho_{\max}}
\!+\!
\frac{\lambda\,\mathbb{E}[S^2]_{\max}}{2(1\!-\!\rho_{\max})^2}
\!\right)\!
\right]
\right. \nonumber \\
&\qquad\left.
+
\frac{\lambda}{b_k(1\!-\!\rho_{\max})}
\!\right\}
\sum_{j=1}^N\!\pi_j c_j .\\[-24pt]\nonumber
\end{align}
In particular, if $L_\infty<1$, then $\widehat{\boldsymbol{\ell}}$ is a contraction on $[0,\ell_{\max}]^N$ in the $\ell_\infty$-norm.
\end{lemma}

\begin{IEEEproof}
By applying the chain rule to \eqref{eq:LK_lambert_fp}, we obtain
\begin{align}
\frac{\partial \widehat{\ell}_k}{\partial \ell_j}
=
\frac{1}{b_k}W'(z_k(\boldsymbol{\ell}))\,\frac{\partial z_k(\boldsymbol{\ell})}{\partial \ell_j}
+\frac{\partial K_k(\boldsymbol{\ell})}{\partial \ell_j},
\label{eq:LW_chain_rule_short}
\end{align}
where $z_k(\boldsymbol{\ell})\!\triangleq\! b_kL_k(\boldsymbol{\ell})e^{-b_kK_k(\boldsymbol{\ell})}$.
For $z>0$, $|W'(z)|\le 1/|z|$. Moreover,
\[
\frac{\partial z_k}{\partial \ell_j}
=
b_k e^{-b_kK_k}\!\left(\!
\frac{\partial L_k}{\partial \ell_j}
-b_kL_k\frac{\partial K_k}{\partial \ell_j}
\!\right)\!.
\]
Using $z_k=b_kL_ke^{-b_kK_k}$ yields
\begin{align}
\!\!\!\left|\frac{\partial \widehat{\ell}_k}{\partial \ell_j}\right| \!\le\!
\frac{1}{b_k|z_k|}\left|\frac{\partial z_k}{\partial \ell_j}\right|
\!+\!\left|\frac{\partial K_k}{\partial \ell_j}\right| \!\le\!
\frac{1}{b_kL_k}\left|\frac{\partial L_k}{\partial \ell_j}\right|
\!+\!2\left|\frac{\partial K_k}{\partial \ell_j}\right|\!.\!\!
\label{eq:LW_basic_bound_short}
\end{align}
From the definitions of $L_k(\boldsymbol{\ell})$ and $K_k(\boldsymbol{\ell})$ and the bounds
$t_j(\ell_j)\le t_{\max}$,
$\mathbb{E}[S^2(\boldsymbol{\ell})]\le \mathbb{E}[S^2]_{\max}$, and
$1-\lambda\mathbb{E}[S(\boldsymbol{\ell})]\ge 1-\rho_{\max}$, we obtain
\begin{align*}
\frac{1}{b_kL_k}\!\left|\!
\frac{\partial L_k}{\partial \ell_j}
\!\right|
&\!\le\!
\frac{\lambda\,\pi_j c_j}{b_k(1\!-\!\rho_{\max})}, \\[2pt]
\left|\!
\frac{\partial K_k}{\partial \ell_j}
\!\right|
&\!\le\!
\frac{\pi_j c_j}{c_k}\!\left[\!
1\!+\!\lambda\!\left(\!
\frac{t_{\max}}{1\!-\!\rho_{\max}}
\!+\!
\frac{\lambda\,\mathbb{E}[S^2]_{\max}}{2(1\!-\!\rho_{\max})^2}
\!\right)\!
\right]\!.
\end{align*}
Substituting these bounds into \eqref{eq:LW_basic_bound_short} proves
\eqref{eq:LW_partial_bound_compact}. Summing over $j$ and taking the maximum
over $k$ yields \eqref{eq:L_infty_LW_compact}, and  $L_\infty\!<\!1$ ensures
contraction by the Banach fixed-point theorem \cite{Banach1922}.\!\!\!
\end{IEEEproof}

\subsection{Gradient Ascent Algorithm}
\label{sec:gd_method}

The projected fixed-point method provided in Subsection~\ref{sec:convergence_fp}
requires that the mapping is contractive. When this condition does not hold,
the iteration may not converge. To ensure convergence for all feasible operating
points, we now consider projected gradient ascent (PGA), which only requires
a bound on how fast the gradient of the objective function $J(\bm{\ell})$ in (\ref{eq:objective}) may change. The PGA iterations are given as
\begin{equation}
\boldsymbol{\ell}^{(n+1)}
=
\mathcal{P}_{[0,\ell_{\max}]^N}
\!\left(
\boldsymbol{\ell}^{(n)} + \eta \nabla J(\boldsymbol{\ell}^{(n)})
\right),
\label{eq:gd_update_v2}
\end{equation}
where $\eta>0$ is the learning rate. Here, $\mathcal{P}_{[0,\ell_{\max}]^N}(.)$ denotes the projection operator onto the feasible set $[0,\ell_{\max}]^N$ applied when the gradient ascent update exceeds this set. Since $J(\bm{\ell})$ is strictly concave on the stability region (due to Lemma~\ref{lem:convexity}), choosing $\eta$ small enough guarantees
convergence to the unique optimum $\boldsymbol{\ell}^\star$ \cite{bertsekas1999nonlinear}. Next, we need to show that the gradient of $J(\bm{\ell})$ is Lipschitz continuous with respect to the $\ell_\infty$-norm if there exists $L_J$ such that we have
\begin{align*}
\|\nabla J(\boldsymbol{x}) - \nabla J(\boldsymbol{y})\|_{\infty}
\le
L_{J} \,\|\boldsymbol{x}-\boldsymbol{y}\|_{\infty},
\quad \forall \boldsymbol{x},\boldsymbol{y},
\end{align*}
which holds whenever $\|\nabla^2 J(\boldsymbol{\boldsymbol{\ell}})\|_{\infty} \leq L_{J}$ for all $\boldsymbol{\ell}$ and for some constant $L_{J}>0$. Then, due to \cite[Prop. 2.3.2]{bertsekas1999nonlinear}, the PGA iteration in (\ref{eq:gd_update_v2}) converges whenever we have 
\begin{align}
0 < \eta < \frac{2}{L_{J}}.
\label{eq:stepsize_rule_v2}
\end{align}
To establish such a constant, we recall that the Lipschitz property of the
gradient is determined by the largest possible second derivative of $J(\bm{\ell})$,
namely by the $\ell_\infty$-norm of its Hessian given by
\[
\max_{\boldsymbol{\ell}\in[0,\ell_{\max}]^N}
\bigl\|\nabla^2 J(\boldsymbol{\ell})\bigr\|_{\infty}.
\]
Since the optimization is restricted by
$0 \le \ell_k \le \ell_{\max}$ and
$\lambda\mathbb{E}[S(\boldsymbol{\ell})] < 1$,
we can compute a uniform bound by evaluating the worst-case values inside this
compact region.

\begin{lemma}
\label{lem:lipschitz_bound}
Let $J(\boldsymbol{\ell})$ be the objective function in~\eqref{eq:objective}. Let us recall the service-time model in Section~\ref{sec:system_model} and also the definitions of $t_k^{\max}$ for $k\in\{1,\cdots,N\}$, $\mathbb{E}[S]_{\max}$, $\mathbb{E}[S^2]_{\max}$, and $\rho_{\max} =  \lambda \mathbb{E}[S]_{\max} < 1$ from Lemma~\ref{lem:contraction_bound_LW}.
For $k,j\in\{1,\dots,N\}$, let us define
\begin{align}
H_{kj}
\triangleq& \frac{\lambda \delta_{kj}\pi_k c_k^2 }{1-\rho_{\max}} + \frac{\lambda^2 \pi_k c_k \pi_j c_j (t_k^{\max}+ t_j^{\max})}{(1-\rho_{\max})^2} \nonumber\\ &+ \frac{\lambda^3 \pi_k c_k \pi_j c_j \mathbb{E}[S^2]_{\max}}{(1-\rho_{\max})^3}+\alpha\delta_{kj}\pi_k A_k b_k^2 ,
\end{align}
where $\delta_{kj}$ is the Kronecker delta defined as $\delta_{kj} = 1$ when $j=k$; and $\delta_{kj} =0$, otherwise. Then, for every feasible
$\boldsymbol{\ell}\!\in\![0,\ell_{\max}]^N$ satisfying
$\!\lambda \mathbb{E}[S(\boldsymbol{\ell})]\! <\! 1$, we have
\[
\left|
\frac{\partial^2 J(\boldsymbol{\ell})}{\partial \ell_k \partial \ell_j}
\right|
\le H_{kj},
\qquad \forall\,k,j,
\]
and consequently
\begin{equation}
\bigl\|\nabla^2 J(\boldsymbol{\ell})\bigr\|_\infty
\;\le\;
L_J
\triangleq
\max_{k}
\sum_{j=1}^N H_{kj},
~
\forall\,\boldsymbol{\ell}\in[0,\ell_{\max}]^N .
\label{eq:HJ_bound_lemma}
\end{equation}
In particular, the gradient $\nabla J$ is globally Lipschitz continuous on the
feasible set with Lipschitz constant $L_J$.
\end{lemma}

\begin{IEEEproof}
We decompose $J(\boldsymbol{\ell})$ into the system time and the accuracy terms, and bound the
second derivatives of each component. First, we find the second derivative of the accuracy term $J(\boldsymbol{\ell})$ in  (\ref{eq:objective}) given as $\alpha \sum_{k=1}^N \pi_k\, p_k(\ell_k) = \alpha \sum_{k=1}^N \pi_k (A_k (1 - e^{-b_k\ell_k}) +D_k)$
\begin{align}
  \!\!\frac{\partial^2 }{\partial \ell_k\partial \ell_j}\! \left(\!\! \alpha\! \sum_{k=1}^N \pi_k\, p_k(\ell_k)\!\!\right) \!\!\! =\!\! \! \begin{cases} 
      \!\!- \alpha\ \!\pi_k A_k b_k^2 e^{-b_k \ell_k}, & \text{\!\!if } j\!=\!k,\!\! \\
      0, & \text{\!\!o.w.\!\!\!\!\!}
   \end{cases}
\end{align}
Then, we can obtain the upper bound for the second derivative of the accuracy term as  $\left|\frac{\partial^2 }{\partial \ell_k^2}\! \left(\!\! \alpha\! \sum_{k=1}^N \pi_k\, p_k(\ell_k)\!\!\right)\right|\leq \alpha \pi_k A_k b_k^2$. This yields the term
$\alpha\,\delta_{kj}\,\pi_k A_k b_k^2$ in $H_{kj}$.

The system time contribution to $J(\boldsymbol{\ell})$ depends on $\boldsymbol{\ell}$ only through
$\mathbb{E}[S(\boldsymbol{\ell})]$ and $\mathbb{E}[S^2(\boldsymbol{\ell})]$. Then, by denoting the system time $S_{sys}=\frac{\lambda\, \mathbb{E}[S^2]}
     {2\left(1 - \lambda\, \mathbb{E}[S]\right)}
+ \mathbb{E}[S] $, we obtain the second derivative of the system time as 
\begin{align}
\frac{\partial^2S_{sys} }{\partial \ell_k \partial \ell_j}
 =& \frac{\lambda \delta_{kj}\pi_k c_k^2 }{1-\lambda \mathbb{E}[S(\boldsymbol{\ell})]} + \frac{\lambda^2 \pi_k c_k \pi_j c_j (t_k(\ell_k)+ t_j(\ell_j))}{(1-\lambda \mathbb{E}[S(\boldsymbol{\ell})])^2} \nonumber\\ &+ \frac{\lambda^3 \pi_k c_k \pi_j c_j \mathbb{E}[S^2(\boldsymbol{\ell})]}{(1-\lambda \mathbb{E}[S(\boldsymbol{\ell})])^3}.   
\end{align}
Then, by using the bounds $t_i(\ell_i) \le t_i^{\max}$ for all $i$, $\mathbb{E}[S(\boldsymbol{\ell})]\le\mathbb{E}[S]_{\max}$, and $\mathbb{E}[S^2(\boldsymbol{\ell})]\le\mathbb{E}[S^2]_{\max}$, and and $\lambda\mathbb{E}[S(\boldsymbol{\ell})]\le \rho_{\max}<1$, we obtain
\begin{align}
\left|\frac{\partial^2S_{sys} }{\partial \ell_k \partial \ell_j}\right|
 \leq & \frac{\lambda \delta_{kj}\pi_k c_k^2 }{1-\rho_{\max}} + \frac{\lambda^2 \pi_k c_k \pi_j c_j (t_k^{\max}+ t_j^{\max})}{(1-\rho_{\max})^2} \nonumber\\ &+ \frac{\lambda^3 \pi_k c_k \pi_j c_j \mathbb{E}[S^2]_{\max}}{(1-\rho_{\max})^3},   
\end{align}
which corresponds to the system time part of $H_{kj}$.

Adding the system time and accuracy contributions yields
\begin{align}
\left|
\frac{\partial^2 J(\boldsymbol{\ell})}{\partial \ell_k \partial \ell_j}
\right|
\le H_{kj},
\qquad \forall\,k,j,
\end{align}
for all feasible $\boldsymbol{\ell}$. Therefore, we have
\begin{align}
\!\!\bigl\|\nabla^2 J(\boldsymbol{\ell})\bigr\|_{\infty}
\!\!=\!
\max_k \sum_{j=1}^N
\left|
\frac{\partial^2 J(\boldsymbol{\ell})}{\partial \ell_k \partial \ell_j}
\right|\!\le\!
L_J
\!=\!
\max_k \sum_{j=1}^N H_{kj},
\end{align}
which establishes~\eqref{eq:HJ_bound_lemma} and the global Lipschitz
continuity of $\nabla J$ with constant $L_J$.
\end{IEEEproof}

Thus, Lemma~\ref{lem:lipschitz_bound} provides the Lipschitz constant
required for the convergence analysis. Substituting the bound
\(
\|\nabla^2 J(\boldsymbol{\ell})\| \le L_J
\)
into the learning rate condition \eqref{eq:stepsize_rule_v2}, we obtain
\begin{equation}
 0 \;< \;\eta \;<\; \frac{2}{L_J} =\frac{2}{\max_{k} \sum_{j=1}^N H_{kj}} ,
\label{eq:eta_final_v2}
\end{equation}

\vspace{4pt}
\noindent
Hence, the PGA update in \eqref{eq:gd_update_v2} converges to the unique optimum
$\boldsymbol{\ell}^\star$ for any feasible initialization, independent of the
contractivity properties of the fixed-point method.

\subsection{Integer Projection}
\label{sec:integer_impl}

All optimization methods described above, the projected fixed-point iteration
in~\eqref{eq:proj_fp_iteration_LK} and the projected gradient ascent
budgets $\{\ell_k\}$ are treated as continuous variables in $[0,\ell_{\max}]$.
Let $\boldsymbol{\ell}^\star$ denote the unique optimal solution of this continuous relaxation. Since the number of reasoning tokens must be an integer in any practical LLM
deployment, we convert $\boldsymbol{\ell}^\star$ into an implementable
integer-valued budget vector by a simple projection onto the discrete set \eqref{eq:integer_tokens}. Let us denote the floor and ceil of the optimal real valued reasoning tokens as $\ell_k^{\mathrm{floor}} = \lfloor \ell_k^*\rfloor$ and $\ell_k^{\mathrm{ceil}} = \lceil \ell_k^* \rceil$, respectively. For each $\ell_k^{\mathrm{int}}$, one can choose between $\ell_k^{\mathrm{floor}}$ and $\ell_k^{\mathrm{ceil}}$, which yields $2^N$ distinct selection combinations. One potential approach is to evaluate all $2^N$ selection combinations and choose the integer solution that minimizes the overall cost function, that is, 
\begin{align}\label{eq:integer_projection_1}
   \bm{\ell}^{\mathrm{int}} = \argmax_{\ell_k \in\{\ell_k^{\mathrm{floor}}, \ell_k^{\mathrm{ceil}}\} } J(\bm{\ell}).
\end{align}
Another convenient choice with lower search complexity is the
component-wise rounding rule, that is,
\begin{equation}
\ell_k^{\mathrm{int,2}}
=\operatorname{round}(\ell_k^\star)
\quad k = 1,\cdots,N,
\label{eq:integer_projection_2}
\end{equation}
where $\operatorname{round}(\cdot)$ denotes rounding to the nearest integer.
By construction, the integer vector
$\boldsymbol{\ell}^{\mathrm{int}}
= (\ell_1^{\mathrm{int}},\dots,\ell_N^{\mathrm{int}})$ (and also $\boldsymbol{\ell}^{\mathrm{int,2}}$)
satisfies the original discrete feasibility constraints
\eqref{eq:integer_tokens} as well as the architectural bounds
$0 \le \ell_k^{\mathrm{int}} \le \ell_{\max}$.

In the following, we derive an optimality gap between the overall optimal solution with the integer-valued reasoning tokens denoted as $\boldsymbol{\ell}^{\mathrm{int}}_{opt}$ and the solution obtained with $\bm{\ell}^{\mathrm{int}} $ obtained in (\ref{eq:integer_projection_1}) (or in (\ref{eq:integer_projection_2})). Note that we have $J(\bm{\ell}^*)\geq J(\boldsymbol{\ell}^{\mathrm{int}}_{opt})\geq J(\bm{\ell}^{\mathrm{int}}) \geq \bar{J}(\bm{\ell}^*)$, that is, the utility with the real-valued reasoning tokens gives the overall upper bound to the optimal utility with the integer-valued reasoning tokens. Then, the utility that we obtained through the rounding operation will be $J(\bm{\ell}^{\mathrm{int}})$ which will give less than the overall optimal integer solution $J(\boldsymbol{\ell}^{\mathrm{int}}_{opt})$. Thus, our goal is to derive an lower-bound to the cost of rounding operation namely $\bar{J}(\bm{\ell}^*)$. For that, with the rounding operation, we obtain integer-valued reasoning tokens $\ell_k^*\!-\!1\!\leq\! \ell_k^{\mathrm{int}}\!\leq\! \ell_k^*\!+\!1$. 
By denoting $c_{\max} = \max_{k}c_k$ and after some algebraic manipulations, we have 
\begin{align}\label{eq:upper_bound_2}
J(\bm{\ell}^*)
\geq \bar{J}(\bm{\ell}^*)=&
\alpha \sum_{k=1}^N \pi_k\, (A_k (1 - e^{-b_k (\ell_k-1)}) + D_{k} ) \nonumber \\
&-\frac{\lambda\mathbb{E}[S^2] + 2c_{\max} }
     {2\left(1 - \lambda (\mathbb{E}[S]+c_{\max})  \right)} -\mathbb{E}[S].
\end{align}
This lower-bound is valid under the assumption that $ \lambda (\mathbb{E}[S]+c_{\max}) <1$ which typically holds since the $c_k$ values are usually small. Therefore, the utility with the integer-valued reasoning tokens will be upper-bounded by $J(\bm{\ell}^*)$ and lower-bounded by $\bar{J}(\bm{\ell}^*)$.  

In the following section, we present numerical results to validate our theoretical analysis.
\begin{table*}[t]
\centering
\caption{Accuracy and latency model parameters and optimal reasoning-token allocations.}
\label{tab:fit_and_opt}
\vspace{-0.25cm}
\begin{tabular}{lccccccc}
\hline
Dataset & $A_k$ & $b_k$ & $D_k$ & $t_{0k}$ & $c_k$ &  $\ell_k^\star$ \\
\hline
AIME          & 0.6808 & $1.59\!\times\!10^{-4}$ & 0     & 0.1380 & 0.0120 &  0.0 \\
GSM8K         & 0.7230 & $3.20\!\times\!10^{-3}$ & 0.277 & 0.1459 & 0.0141 &  340.5 \\
GPQA          & 0.3552 & $4.41\!\times\!10^{-4}$ & 0.276 & 0.1674 & 0.0126 &  0.0 \\
CRUXEval      & 0.4379 & $5.63\!\times\!10^{-4}$ & 0     & 0.0176 & 0.0124 &  0.0 \\
BBH           & 0.7146 & $1.75\!\times\!10^{-3}$ & 0.148 & 0.2073 & 0.0127 & 345.0 \\
ARC-Challenge & 0.3933 & $1.66\!\times\!10^{-1}$ & 0.490 & 0.0581 & 0.0119 &  30.1 \\
\hline
\end{tabular}
\vspace{-0.5cm}
\end{table*}
\section{Numerical Results}
\label{sec:numerical_results}
For the simulation results, we consider $N=6$ heterogeneous task types corresponding to the datasets
\emph{AIME} \cite{hendrycks2021math}, \emph{GSM8K} \cite{cobbe2021gsm8k}, \emph{GPQA} \cite{rein2023gpqa}, \emph{CRUXEval} \cite{gu2024cruxeval}, \emph{BBH} \cite{suzgun2022bbh}, and
\emph{ARC-Challenge} \cite{clark2018arc}.
All experiments use the \text{Qwen3-8B} \cite{qwen3_report} model and are executed on Google Colab
with an NVIDIA A100 (80\,GB) GPU. The total arrival rate is $\lambda=0.1$ and the importance weight 
for the accuracy in \eqref{eq:objective} is chosen as $\alpha=30$. We set the maximum reasoning-token budget to $\ell_{\max}=32768$, which corresponds
to the maximum number of reasoning tokens supported by the \text{Qwen3-8B} model \cite{qwen3_report}.
We assume a uniform task mixture, i.e., $\pi_k = 1/6$ for all $k$. To simulate the queueing dynamics under FIFO service, we generate a stream of 10{,}000 questions according to the specified Poisson arrival process and task mixture.
Throughout, we keep the inference configuration fixed and set the generation
temperature to $0.1$, while varying only the per-type reasoning-token budgets
$\{\ell_k\}$.

\vspace{-0.1cm}
\subsection{Model Fit and Optimal Budgets}
\vspace{-0.1cm}
For each dataset and each reasoning-token budget, we evaluate on 250
instances and report results averaged over 3 independent runs.
We record empirical accuracy and end-to-end latency
under the enforced reasoning-token budget.
These measurements are used to fit the accuracy and service-time models
introduced in Section~\ref{sec:system_model}.
Table~\ref{tab:fit_and_opt} reports the fitted accuracy and latency parameters
together with the optimal reasoning-token allocation $\boldsymbol{\ell}^\star$
obtained by solving \eqref{eq:opt_problem}. We also plot the fitted accuracy graphs in Fig.~\ref{fig:accuracy_curve}. 
The resulting allocation exhibits pronounced heterogeneity across tasks.
In particular, \emph{GSM8K} and \emph{BBH} receive the largest budgets, indicating
substantial marginal benefit from additional reasoning as it can be also seen in Fig.~\ref{fig:accuracy_curve}.
In contrast, \emph{AIME}, \emph{GPQA}, and \emph{CRUXEval} are assigned zero tokens. Here, \emph{AIME}'s and \emph{CRUXEval}'s accuracy gains are not sufficient to offset the induced
system-time penalty under the operating point $(\lambda,\alpha)$. On the other hand, for \emph{GPQA}, the accuracy of the LLM server with no reasoning token allocation is larger than 0 and improves slowly with the number of reasoning tokens. For that, \emph{GPQA} also receives zero token allocation. Finally, \emph{ARC-Challenge} receives a small but non-zero allocation, consistent with a rapidly saturating improvement profile.
\begin{figure}[t]
\centering
\includegraphics[width=0.9\columnwidth]{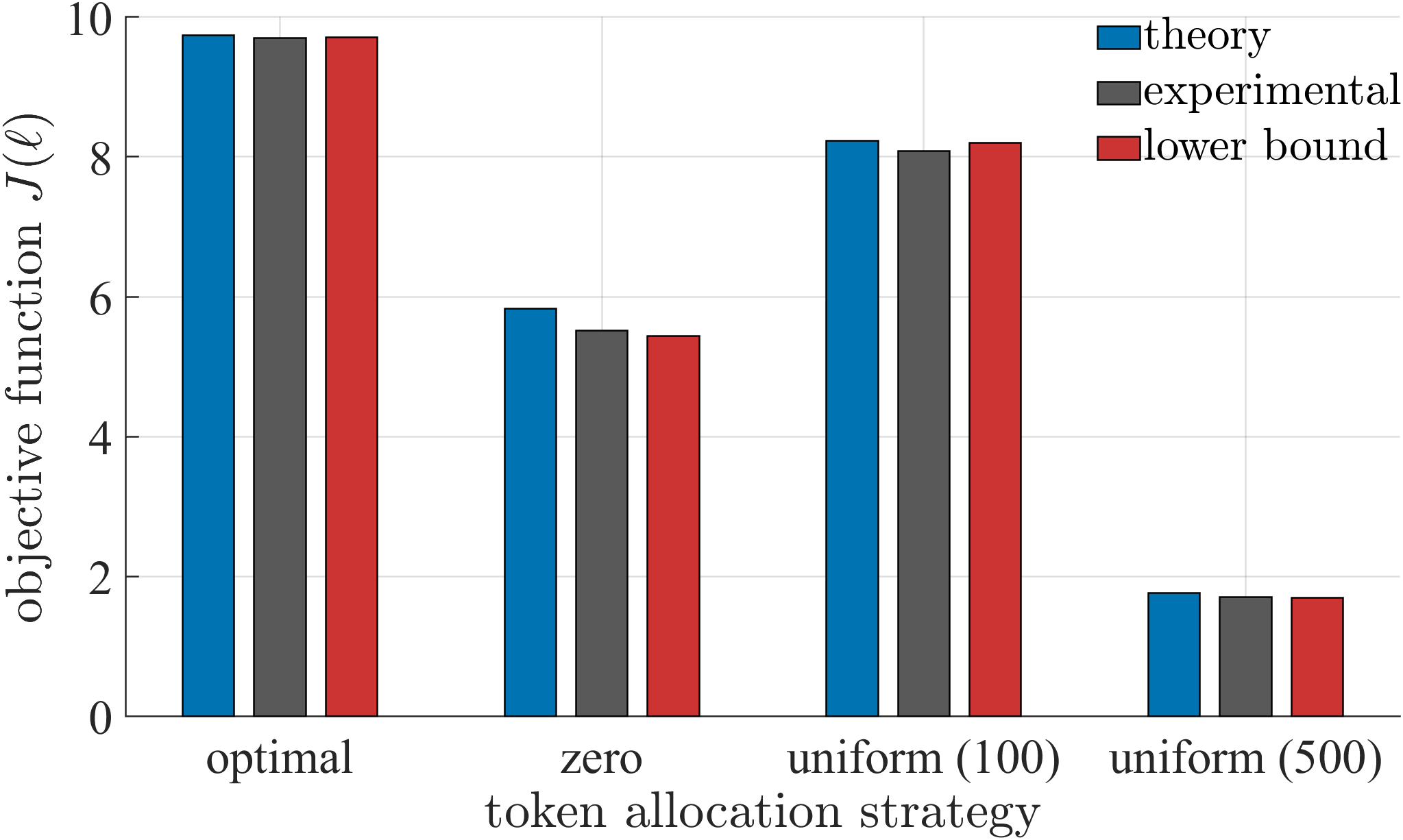}
\vspace{-0.25cm}
\caption{Objective value $J(\boldsymbol{\ell})$ under uniform token allocations
($\ell_k \in \{0,100,500\}$) and the proposed optimal heterogeneous allocation
$\boldsymbol{\ell}^\star$.}
\label{fig:J_bar}
\vspace{-0.5cm}
\end{figure}
\vspace{-0.05cm}
\subsection{Performance Under Alternative Allocation Policies}
\vspace{-0.05cm}
We compare the achieved objective value $J(\boldsymbol{\ell})$ under
(i) uniform token allocation with $\ell_k \in \{0,100,500\}$ for all tasks, and
(ii) the optimal heterogeneous allocation $\boldsymbol{\ell}^\star$. As shown in Fig.~\ref{fig:J_bar}, the proposed allocation yields the best system-level
utility among all compared policies.
Uniformly allocating a large budget (e.g., $\ell_k=500$) leads to excessive queueing
delay and significantly degrades the objective, whereas assigning no reasoning tokens
sacrifices accuracy gains that are beneficial for certain tasks.
The optimal policy resolves this trade-off by concentrating reasoning effort on
tasks with high utility while avoiding latency inflation on tasks with
limited gains.

\vspace{-0.05cm}
\subsection{Sensitivity to the GSM8K Budget}
\begin{figure}[t]
\centering
\includegraphics[width=0.9\columnwidth]{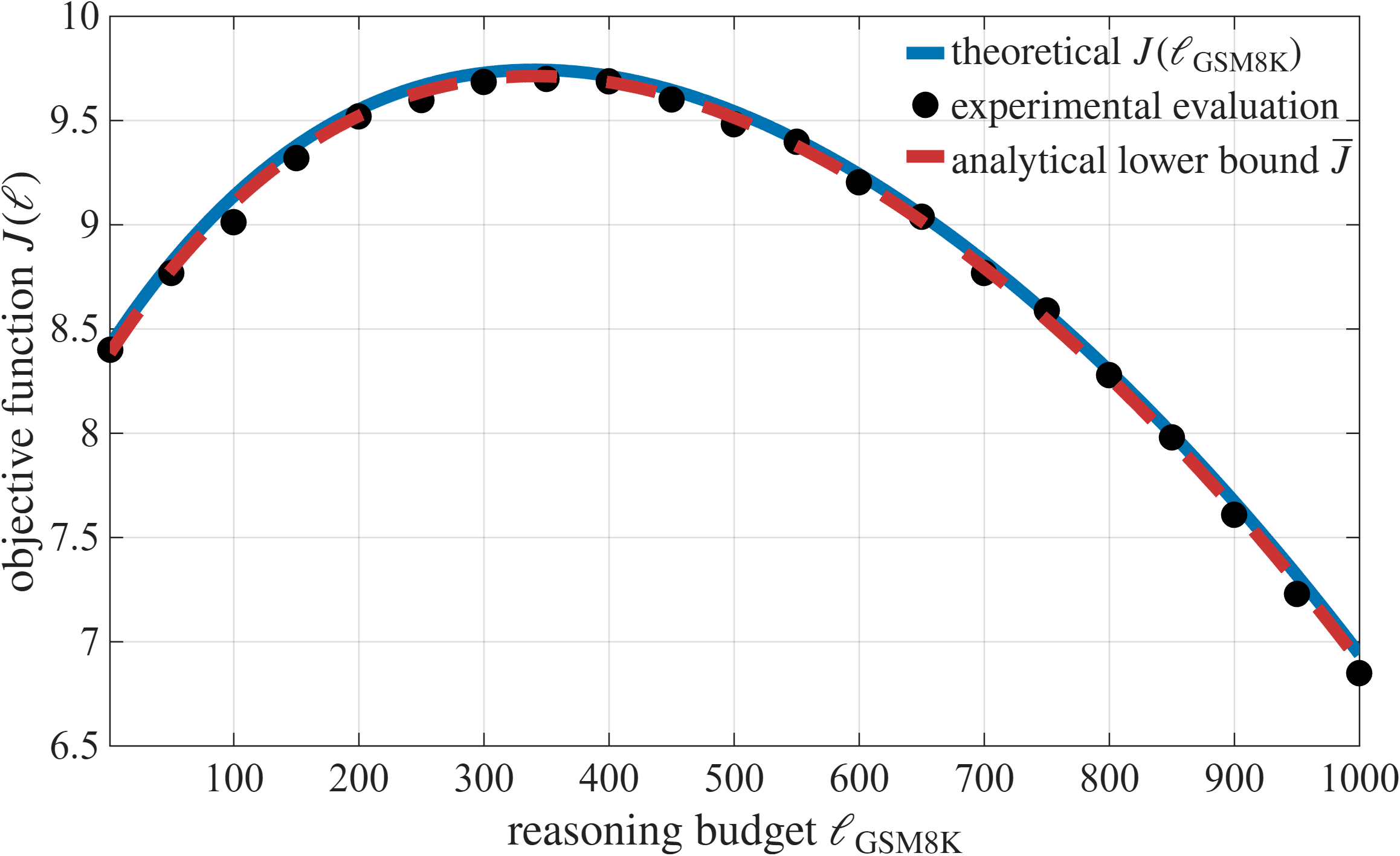}
\vspace{-0.25cm}
\caption{Objective value $J(\boldsymbol{\ell})$ as a function of the GSM8K reasoning-token budget
$\ell_{\mathrm{GSM8K}}$, with all other budgets fixed to their optimal values.}
\label{fig:GSM8K_sweep}
\vspace{-0.5cm}
\end{figure}

\vspace{-0.05cm}
To assess sensitivity and validate the predicted trade-off structure, we fix all
reasoning-token budgets except $\ell_{\mathrm{GSM8K}}$ to their optimal values and
vary $\ell_{\mathrm{GSM8K}}$ from $0$ to $1000$. Fig.~\ref{fig:GSM8K_sweep} shows the resulting dependence of the objective value
$J(\boldsymbol{\ell})$ on $\ell_{\mathrm{GSM8K}}$.
The curve exhibits an unimodal structure, with a unique maximizer attained
at $\ell_{\mathrm{GSM8K}}\approx 340$, which closely matches the optimal value
reported in Table~\ref{tab:fit_and_opt}.
This confirms that the optimal token budget identified by the
theoretical optimization is consistent with empirical observations. Moreover, Fig.~\ref{fig:GSM8K_sweep} illustrates the underlying accuracy--latency trade-off: for small token budgets, increasing $\ell_{\mathrm{GSM8K}}$ yields significant
accuracy improvements that dominate the additional delay, leading to an increase
in $J(\bm\ell)$.
Beyond the optimal point, however, the accuracy gains saturate while the induced
queueing delay continues to grow, causing a monotonic decrease in the overall
system utility. 

In Fig.~\ref{fig:GSM8K_sweep}, the blue curve represents the server utility evaluated at the optimal real-valued reasoning-token allocation, $J(\boldsymbol{\ell}^*)$, while the red dashed curve denotes the lower bound on the utility, $\bar{J}(\boldsymbol{\ell}^*)$, derived in Subsection~\ref{sec:integer_impl}. We observe that the experimental results, shown by the black circles, closely track this theoretical bound; however, in some instances they slightly exceed these bounds due to sample variance. As the number of samples increases, the experimental results are expected to lie between the derived theoretical bounds. 

\vspace{-0.1cm}
\section{Conclusion}
\vspace{-0.1cm}
We studied token-budget allocation for heterogeneous LLM queries served by a single FIFO $\mathrm{M}/\mathrm{G}/1$ system, capturing the joint accuracy-latency trade-off induced by reasoning tokens. By formulating a utility that rewards weighted accuracy and penalizes mean system time, we proved strict concavity over the stability region, which guarantees a unique optimal allocation. We then developed projected fixed-point iterations with explicit contraction conditions and a projected gradient method with a global step-size bound to ensure convergence more broadly. Finally, we proposed practical integer token budgets via rounding and characterized the associated performance loss. Our results show that heterogeneous, task-aware token allocation can substantially improve system-level performance compared to uniform budgeting.

\bibliographystyle{IEEEtran}
\bibliography{references}

\end{document}